\documentclass[conference]{IEEEtran}
\usepackage{times}

\usepackage[numbers]{natbib}
\usepackage{multicol}
\usepackage[bookmarks=true]{hyperref}
\usepackage{subfigure}
\usepackage{graphicx}
\usepackage{amsmath}
\usepackage{amssymb}
\usepackage{algorithm}
\usepackage[noend]{algpseudocode}
\usepackage{color}
\usepackage{colortbl}
\usepackage{tabularx}
\usepackage{authblk}

\newcommand{\cmt}[1]{}

\long\def\ignorethis#1{}

\newcommand{\etal}{{\em{et~al.}\ }}
\newcommand{\eg}{e.g.\ }
\newcommand{\ie}{i.e.\ }

\newcommand{\vc}[1]{\ensuremath{\mathbf{#1}}}

\newcommand{\argmin}{\operatornamewithlimits{argmin}}

\newcommand{\pctab}{\hspace{0.2in}}

\begin{document}

\title{Preparing for the Unknown: Learning a Universal Policy with Online System Identification}

\author[1]{Wenhao Yu}
\author[2]{Jie Tan}
\author[1]{C. Karen Liu}
\author[1]{Greg Turk}
\affil[ ]{\textit{wenhaoyu@gatech.edu, jietan@google.com, karenliu@cc.gatech.edu, turk@cc.gatech.edu}}
\affil[1]{Interactive Computing, Georgia Institute of Technology, USA}
\affil[2]{Google Brain, Google, USA}
\maketitle

\begin{abstract}

We present a new method of learning control policies that successfully operate under unknown dynamic models.  We create such policies by leveraging a large number of training examples that are generated using a physical simulator.  Our system is made of two components: a Universal Policy (UP) and a function for Online System Identification (OSI).  We describe our control policy as “universal” because it is trained over a wide array of dynamic models.  These variations in the dynamic model may include differences in mass and inertia of the robot’s components, variable friction coefficients, or unknown mass of an object to be manipulated.  By training the Universal Policy with this variation, the control policy is prepared for a wider array of possible conditions when executed in an unknown environment.  The second part of our system uses the recent state and action history of the system to predict the dynamics model parameters $\boldsymbol{\mu}$.  The value of $\boldsymbol{\mu}$ from the Online System Identification is then provided as input to the control policy (along with the system state).  Together, UP-OSI is a robust control policy that can be used across a wide range of dynamic models, and that is also responsive to sudden changes in the environment.  We have evaluated the performance of this system on a variety of tasks, including the problem of cart-pole swing-up, the double inverted pendulum, locomotion of a hopper, and block-throwing of a manipulator. UP-OSI is effective at these tasks across a wide range of dynamic models. Moreover, when tested with dynamic models outside of the training range, UP-OSI outperforms the Universal Policy alone, even when UP is given the actual value of the model dynamics. In addition to the benefits of creating more robust controllers, UP-OSI also holds out promise of narrowing the Reality Gap between simulated and real physical systems.

\end{abstract}

\IEEEpeerreviewmaketitle

\section{Introduction}


Numerical simulation of physical phenomena is a powerful tool that has been embraced by researchers and practitioners in computer animation. The success in simulating highly dynamic motion in computer animation, however, has not been transferred in full to robotics. The discrepancy between what can be achieved in simulation and that in real world is referred as the ``Reality Gap'' in the Evolutionary Robotics community \cite{realityGap,conf/gecco/KoosMD10}. Researchers have put forth a long list of possible factors that give rise to the Reality Gap, such as simplified dynamic models, inaccurate model parameters, approximated hardware limitations, the absence of uncertainty and latency in sensors and actuators, and other unmodelled factors. Closing the Reality Gap has recently been a major focus in robotics because the ability to transfer knowledge learned in simulation to the real world can potentially unlock the full capability of deep reinforcement learning for robotic applications.

There are two general approaches to learning control policies for real-world operations. One can design a robust control policy to handle a reasonable amount of noise based on a hypothesized dynamic model and hope that the control policy will succeed in the real world without any intervention. This approach is effective if the hypothesized dynamic model is not too far off from the real one.  Unfortunately, this is typically not the case for complex robotic system that are performing dynamic tasks involving contacts. Alternatively, one can learn a more accurate dynamic model from real-world data by alternating between control policy optimization, data collection, and dynamic model fitting (or \emph{system identification}). One potential limiting factor of this approach is the need for a large amount of real-world trials which can be expensive, time-consuming, and sometimes dangerous to the robot,  humans and objects in its surroundings.

This paper introduces a different approach to combat the Reality Gap. We present a new algorithm that only uses data from a generic physical simulator to learn motor control policies that successfully operate under unknown dynamic models. Our key idea is to aggressively explore  the ``virtual world'' through physical simulation and precompute many of the possible situations the robot might encounter when operating in real world. Although it is easy to dismiss such an approach based on the curse of dimensionality argument, in light of recent disruptive development in deep learning and the availability of large-scale computing capability, we believe it is time to revisit this pre-computation approach. Suppose that we have a way to precompute the optimal control policy for every dynamic model in a parameterized space. Suppose further that we have a fast method that tells us which dynamic model best fits an observed motion sequence. With these two scenarios, during online execution in the real world, we only need to select the right dynamic model by looking at the recent history of robot motion, and then select the corresponding control policy for that dynamic model to achieve optimal motion. While it might not be possible to pre-learn a control policy or pre-explore the entire space of motions for every possible dynamic model, it might be possible to do so for everything that is important for the task of interest.

To this end, we introduce a new approach that exploits a massive amount of simulated data to learn a) a universal control policy (UP) that is capable under a parameterized space of dynamic models, when provided with the appropriate dynamic model parameters, and b) an on-line system identification model (OSI) that predicts the dynamic model parameters given the current state and the recent history of state-action pairs. Figure \ref{fig:contributions} shows the relationship between the universal policy and the on-line system identification. Once trained, the combined algorithm, UP-OSI, can be executed in an unknown dynamic environment. At each time instance, the dynamic model parameters $\boldsymbol{\mu}$ is first predicted by the learned system identification model. The universal control policy then takes the predicted model parameters along with the current state to compute the the optimal action (Figure \ref{fig:contributions}).

UP-OSI is sample-efficient by design because the algorithm does not require real-world samples during offline training. Another important advantage of UP-OSI is that it does not require the model parameters to be identified prior to execution. While some model parameters might not change over time (\eg mass, length of a body part) and can be identified offline, other parameters related to the unknown environment, such as the friction coefficient of the floor or the mass of objects being manipulated, cannot be easily identified in advance. Part of the power of UP-OSI is that it can dynamically adapt to changing factors in the environment.

We evaluate our method by learning dynamic motor skills and executing them under unknown dynamic models in simulation. In each of the examples, the control policy can successfully execute the task without knowing some crucial parameters of the dynamic model, such as the inertial and geometric parameters of the robot, variable friction coefficients in the environment, and other task-related parameters. Furthermore, we demonstrate that UP-OSI can operate successfully outside the space of dynamic models used for training, as well as under sudden changes in the environment.

\section{RELATED WORK}


\subsection{Deep Reinforcement Learning}
In recent years, researchers have used deep reinforcement learning to train highly dynamic motor skills in simulated environments that have high-dimensional state and action spaces ~\cite{2016-TOG-deepRL, schulman2015high, schulman2015trust, lillicrap2015continuous, mnih2016asynchronous, gu2016q}. For example, Schulman \etal demonstrated learning of fullbody humanoid running and getting up with just feedback from the reward function using Trust Region Policy Optimization (TRPO) \cite{schulman2015trust} and Generalized Advantage Estimation (GAE) \cite{schulman2015high}. \citet{lillicrap2015continuous} extended their work of Deep Q-Learning~\cite{mnih2015human} and Deterministic Policy Gradient (DPG)~\cite{Silver2014DeterministicPG} to learn robotic motor skills such as hopping, reaching and 2D walking directly from pixel input. These methods usually require a large amount of interaction time between the agent and the environment, which poses a significant challenge to the robot as well as the experimenter when applying them directly to learn a real-world robotic control task.


In addition, progress has been made in directly learning neural network control policies of manipulation tasks for real robots~\cite{DBLP:journals/corr/LevineFDA15, pinto2016supersizing,DBLP:journals/corr/LevinePKQ16}. While the results are impressive, these methods usually require extensive amount of experimental data \cite{DBLP:journals/corr/LevinePKQ16,pinto2016supersizing} or  relatively restrictive settings \cite{DBLP:journals/corr/LevineFDA15}. It is unclear whether these method would work directly on more dynamic motor skills in the real-world, such as locomotion.


\subsection{Transfer Learning in Reinforcement Learning}

Transferring policy learned in simulation to real-world robot has the potential to address the problem of learning complex motor skills for real robots. Much previous effort focused on classical \emph{system identification}, which provides a framework to address the general problem of model inconsistency. In practice, system identification is often interleaved with control policy optimization to minimize the number of required real-world experiments \cite{Abbeel:2005,gevers2006,journals/tec/BongardL05}. Some widely-used models, such as linear models \cite{Abbeel:2005}, Gaussian processes \cite{deisenroth2011,HA:2015}, and differential equations \cite{zagal2004,Abbeel:2006}, have proven effective for continuous dynamics and control tasks with relatively low action space. For example, \citet{Abbeel:2005} used a time-variant linear function to model the dynamics of a helicopter from the real-world data while learning a control policy to perform inverted autonomous helicopter flight. \citet{deisenroth2011} trained a Gaussian process model from real-world data to analytically calculate the control policy gradient, significantly reducing the number of samples compared to sampling-based policy gradient estimation. \citet{Ross12agnosticsystem} provided a proof that such iterative processes can converge to an optimal policy, given an accurate dynamic model learning method and a good policy search algorithm. More recently, deep neural networks have been applied to learn both forward dynamics \cite{punjani2015deep} and inverse dynamics \cite{christiano2016transfer} from the real-world data, which can potentially model more complicated dynamics. A key drawback of these methods is that their success depends on the quantity and the quality of the real-world data. For highly dynamic or contact-rich tasks, learning an accurate dynamic model and control policy would require a large amount of high quality data, which can be difficult to acquire.

Another line of research utilizes simulated data to train a policy and then directly applies it or adapts it to the real world. \citet{james20163d} demonstrated a simulation-trained manipulation controller learned from photo-realistic rendering data that show similar behaviors in the real-world. \citet{rusu2016sim} used progressive networks to efficiently learn a manipulation task on a Jaco arm from a policy that was trained in simulation. One important assumption made by this approach is that the dynamics modeled by the simulator is similar to the real world dynamics. It is unclear whether these methods can work in the situations where the policy is sensitive to the discrepancy in the assumed dynamics model.


In this work, we avoid the need of explicit modeling of system dynamics by learning a Universal Policy (UP) that simultaneously optimizes a wide range of model parameters and an Online System Identification (OSI) network that estimates the dynamic parameters during execution. The concept of UP is similar to \citet{Mordatch:2015} who optimized the motion trajectory for an ensemble of dynamic models perturbed from the assumed one. Their method shows high success rate when tracking a real-world reference trajectory. The motor skills we are studying require a complex feedback policy, which cannot be achieved by tracking a reference trajectory. \citet{rajeswaran2016epopt} also proposed training a control policy using an ensemble of dynamic models. They used an adversarial training scheme to improve the policy performance while iteratively updating the distribution of the dynamic models using data from the target environment. The resulting controller is robust to unknown parameters. In contrast, training our UP to achieve a high reward for a variety of dynamic models is an easier learning task because UP explicitly takes model parameters as part of the input.

\subsection{Learning Policy in Unknown Environment}

Our work is also related to learning a control policy in an environment with unknown parameters, \ie Partial Observable Markov Decision Process (POMDP). One example of such work is the idea of event-learning proposed by \citet{szita2002varepsilon}. They demonstrated that when combined with Static and Dynamic State (SDS) controller, their method performs better than standard method like SARSA in an dynamically varying environment. In contrast to their work, we explicitly incorporate the model parameters as the input to the control policy. With this additional information as input, the control policy has the potential to achieve better performance for a larger range of model parameters.

Based on the Deterministic Policy Gradient method \cite{Silver2014DeterministicPG}, \citet{heess2015memory} proposed a method that represents a control policy as a recurrent neural network. They demonstrated control policies that identify and memorize task-related signals presented in the observable data sequences, such as the mass of the robot or the position of the target to be grasped. Our work shares a similar goal in learning controllers that can be applied to systems with unknown dynamic model parameters. However, their focus is on learning memory-related tasks rather than transfer to unseen dynamic environments. On the other hand, our work currently does not handle memory-related tasks as our controller only takes a small window of motion history as input.

\section{Methods}

\begin{figure}[t!]
\centering
\subfigure{\includegraphics[width=\columnwidth]{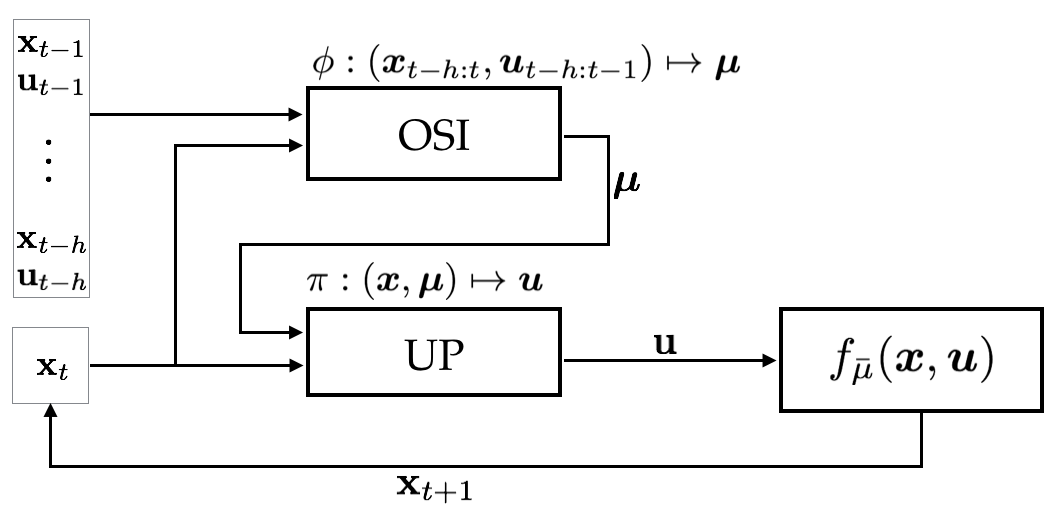}}
\vspace{-4mm}
\caption{Overview of UP-OSI. The online system identification model (OSI) takes as input the recent history of the motion and identify the model parameters $\boldsymbol{\mu}$. The universal control policy (UP) then takes the predicted model parameters along with the current state $\vc{x}$ to compute the optimal control $\vc{u}$.}
\label{fig:contributions}
\vspace{-4mm}
\end{figure}

Our algorithm consists of two components, a universal policy (UP) and an on-line system identification model (OSI), shown in Figure \ref{fig:contributions}. First, we formulate a reinforcement learning problem to learn a universal policy (UP), $\pi: (\vc{x}, \boldsymbol{\mu}) \mapsto \vc{u}$, for a space of dynamic models, $\vc{x}_{t+1} = f_{\boldsymbol{\mu}}(\vc{x}_t,\vc{u}_t)$, parameterized by the dynamic model parameters $\boldsymbol{\mu}$. Unlike a conventional control policy which maps a state vector $\vc{x}$ to a control vector $\vc{u}$, UP takes as input both the state and the dynamic model parameters (\ie $\boldsymbol{\mu}$) and outputs a control vector. Second, we formulate a supervised learning problem to train an online system identification model (OSI), $\phi: (\vc{x}_{t-h:t}, \vc{u}_{t-h:t-1}) \mapsto \boldsymbol{\mu}$, that predicts the dynamic model parameters $\boldsymbol{\mu}$, given the current state $\vc{x}_t$ and the past $h$ time instances of the state-action pairs. Unlike the conventional system identification approach, our goal is not to identify a particular system, but to create a \emph{function} that can identify the model parameters for any given trajectory. Both components are represented as a standard neural network and trained offline using simulated data only.

Putting UP and OSI together, at every time instance, we first use OSI ($\phi$) to predict the dynamic model parameters $\boldsymbol{\mu}$ based on the current state of the robot $\vc{x}_t$ and the recent history of motion $(\vc{x}_{t-1}, \vc{u}_{t-1}, \cdots, \vc{x}_{t-h}, \vc{u}_{t-h})$. Once $\boldsymbol{\mu}$ is identified, we feed both $\boldsymbol{\mu}$ and the current state $\vc{x}_t$ into UP ($\pi$) to evaluate the the optimal action $\vc{u}_t$ under the predicted dynamic model. We execute $\vc{u}_t$ on the robot and push $\vc{x}_t$ and $\vc{u}_t$ into the history queue. The new state of the system becomes the current state $\vc{x}_t$ and the algorithm advances to the next time step (Fig \ref{fig:contributions}).

We propose a novel framework to train the control policy and the system identification model. Conventional approaches alternate between system identification and control policy optimization, which requires a reasonable initial policy or/and an initial system identification model to optimize efficiently. Our method breaks this mutual dependence by first training UP preemptively to anticipate all of the possible dynamic models that OSI might explore during optimization. 

\vspace{-1mm}
\subsection{Learning Universal Policy}
\label{ssec:UCP}
Our goal is to learn a control policy that can be generalized to a parameterized space of dynamic models. Many existing methods \cite{da2012learning, stulp2013learning} employ an ensemble approach by learning a discrete set of control policies and consolidating them into one regression model. Our initial attempt with the ensemble approach showed that, for many dynamic tasks, sometimes a small change in the model parameter requires a drastically different control policy to succeed at the given task. Fitting a regression model to this non-smooth landscape of policies often yields poor generalizability.


In this work, we found that it is possible to directly train a large neural net to represent a universal control policy, $\pi(\vc{x}, \boldsymbol{\mu})$, for a space of dynamic models parameterized by $\boldsymbol{\mu}$. With a powerful policy optimization algorithm and sufficient data, the universal policy can achieve high rewards across the space of $\boldsymbol{\mu}$, with comparable performance to policies that have been trained for a specific $\boldsymbol{\mu}$. 

We use the Trust Region Policy Optimization (TRPO) method \cite{schulman2015trust} and show that by simply augmenting the input state with the model parameters $\boldsymbol{\mu}$, TRPO can successfully train a universal control policy. However, we need to modify the exploration scheme of TRPO because the part of the state space that represents $\boldsymbol{\mu}$ is not affected by forward simulation when generating rollouts. Our algorithm (Algorithm 1) samples $K$ different $\boldsymbol{\mu}$ from a uniform distribution $\rho_\mu$. For each $\boldsymbol{\mu}_i$, we generate a set of rollouts under the policy $\pi(\vc{x}, \boldsymbol{\mu}_i)$ and the dynamic model $f_{\boldsymbol{\mu}_i}$. Once the state-action pairs are collected in this manner, the update of $\pi$ follows TRPO exactly.



\subsection{Learning Online System Identification Model}
Even with the ability to perform control under different dynamic models, UP can only succeed at a task when given accurate model parameters, and this information is typically not readily available. We propose to learn an online system identification model (OSI), $\phi: (\vc{x}_{t-h:t}, \vc{u}_{t-h:t-1}) \mapsto \boldsymbol{\mu}$, that continuously identifies the correct model parameters $\boldsymbol{\mu}$ for UP, when given a short recent history of the states and actions.

The training process can be formulated as a supervised learning problem with the input being a history rollout $H$ and the output being the model parameters $\boldsymbol{\mu}$ under which the input rollout is generated:

\begin{equation}
\theta^* = \argmin_{\theta} \sum_{(H_i, \boldsymbol{\mu}_i) \subseteq B} \|\phi_\theta(H_i) - \boldsymbol{\mu}_i\|^2
\label{eqn:supervise}
\end{equation}
where $\theta$ are the parameters of the neural net $\phi_\theta$. 

Although the training data can be entirely obtained from simulation, the amount of data can be intractably large to thoroughly cover the input space. Our key observation is that OSI only needs to be accurate for the trajectories that are likely to be observed when performing the tasks of interest. As such, we randomly sample the space of $\boldsymbol{\mu}$ where UP is trained for. For each sampled $\boldsymbol{\bar{\mu}}_i$, we simulate $N$ rollouts using the policy $\pi(\vc{x}, \boldsymbol{\bar{\mu}}_i)$ and under the dynamics $f_{\boldsymbol{\bar{\mu}_i}}$. We then generate short history segments from each rollout and store them in the training buffer $B$ (Line 3-13, Algorithm 2).

After optimizing $\phi$ using Equation \ref{eqn:supervise}, we found that the performance of the combined system, UP-OSI, was much worse than simply using UP given true model parameters $\boldsymbol{\bar{\mu}}$. This result is not surprising (retrospectively) because our OSI has only ``seen'' the motion sequences generated by a control policy $\pi(\vc{x}, \boldsymbol{\bar{\mu}})$ under a dynamic model $f_{\boldsymbol{\bar{\mu}}}$ where their model parameters are consistent. In other words, all the training examples so far only cover the ``good cases'' where the control policy is operating optimally under a given dynamic model. When we tested OSI with an unseen initial sequence, OSI was likely to make some error in the prediction. This error is exacerbated because the next sequence that OSI will see is generated by a control using an erroneously predicted $\boldsymbol{\hat{\mu}}$ under the true model parameters $f_{\boldsymbol{\bar{\mu}}}$, where $\boldsymbol{\hat{\mu}} \neq \boldsymbol{\bar{\mu}}$.

Our solution is to iteratively improve OSI by introducing ``bad cases'' with mismatched model parameters used for control ($\pi(\vc{x}, \boldsymbol{\hat{\mu}}$)) and for forward simulation ($f_{\boldsymbol{\bar{\mu}}}$). For each iteration, we generate more training examples using the current OSI and UP. We randomly sample in the space of $\boldsymbol{\mu}$ and generate rollouts like before. However, we feed the $\boldsymbol{\mu}$ predicted by the current OSI into UP, instead of the true model parameters $\boldsymbol{\bar{\mu}}$ used for forward simulation. Note that in Line 25 of Algorithm 2, the dynamic model has the parameter $\boldsymbol{\bar{\mu}}$ which is different from the one used for the control (Line 22, 24). Mixing the mismatched training examples with previously generated ones, we train OSI again using Equation \ref{eqn:supervise}. After a small number of iterations (3-5, see Section IV), the performance of UP-OSI becomes close to the performance of UP that is provided with the true model parameters.

\begin{algorithm}[t]
\caption{Learning UP}\label{alg:learning_alg}
\begin{algorithmic}[1]
\State Randomly initialize UP network $\pi$ \;
\For{$i=1:K$}
\State Initialize rollout buffer $R$ \;
\State $\boldsymbol{\mu} \sim \rho_\mu$ \;
\State $\vc{x} \sim \rho_0$ \;
\While{$R.size \leq MaxStep$}
\State $\vc{u} = \pi(\vc{x}, \boldsymbol{\mu})$ \;
\State $\vc{x} = f_{\boldsymbol{\mu}}(\vc{x}, \vc{u})$ \;
\State $r, terminated = Reward(\vc{x}, \vc{u})$
\State Push $(\vc{x}, \vc{u}, r)$ into $R$ \;
\If{$terminated$}
\State $\boldsymbol{\mu} \sim \rho_\mu$ \;
\State $\vc{x} \sim \rho_0$ \;
\EndIf
\EndWhile
\State Update $\pi$ with data in $R$ using TRPO \;
\EndFor
\Return{$\pi$}
\end{algorithmic}
\end{algorithm}

\begin{algorithm}
\caption{Learning OSI}\label{alg:learning_alg}
\begin{algorithmic}[1]
\State Randomly initialize OSI network $\phi$ \;
\State Initialize training buffer $B$ \;
\For{$i=1:K$}
\State $\boldsymbol{\bar{\mu}} \sim \rho_\mu$ \;
\For{$j=1:N$}
\State Initialize history queue $H$ \;
\State Fill $H$ by simulating under $\pi(\vc{x}, \boldsymbol{\bar{\mu}})$ and $f_{\boldsymbol{\bar{\mu}}}$\;
\For{$t=0:T-1$}
\State Pop $H$ \;
\State $\vc{u}_t = \pi(\vc{x}_t, \boldsymbol{\bar{\mu}})$ \;
\State $\vc{x}_{t+1} = f_{\boldsymbol{\bar{\mu}}}(\vc{x}_t, \vc{u}_t)$ \;
\State Push $(\vc{x}_{t+1}, \vc{u}_t)$ in $H$ \;
\State Store $(H, \boldsymbol{\bar{\mu}})$ in $B$ \;
\EndFor
\EndFor
\EndFor
\State Optimize $\phi$ using data in $B$ \;
\While{not converge}
\For{$i=1:K$}
\State $\boldsymbol{\bar{\mu}} \sim \rho_\mu$ \;
\For{$j=1:N$}
\State Initialize history queue $H$ \;
\State Fill $H$ by simulating under $\pi(\vc{x}, \boldsymbol{\bar{\mu}})$ and $f_{\boldsymbol{\bar{\mu}}}$\;
\For{$t=0:T-1$}
\State $\boldsymbol{\hat{\mu}} = \phi(H)$ \;
\State Pop $H$ \;
\State $\vc{u}_t = \pi(\vc{x}_t, \boldsymbol{\hat{\mu}})$ \;
\State $\vc{x}_{t+1} = f_{\boldsymbol{\bar{\mu}}}(\vc{x}_t, \vc{u}_t)$ \;
\State Push $(\vc{x}_{t+1}, \vc{u}_t)$ in $H$ \;
\State Store $(H, \boldsymbol{\bar{\mu}})$ in $B$\;
\EndFor
\EndFor
\EndFor
\State Optimize $\phi$ using data in $B$ \;
\EndWhile
\Return $\phi$
\end{algorithmic}
\end{algorithm}


\section{Evaluation}

We evaluate UP-OSI on four dynamic motor control problems. In each example, the control policy does not know the true model parameters in advance and relies on OSI to identify the parameters during execution. We vary different model parameters, such as mass, inertia, friction coefficient, or task-related parameters to demonstrate that UP-OSI can successfully perform all the motor skills under unknown dynamic models. We compare the performance of UP-OSI against the performance of the condition, \emph{UP-true}, which uses UP given the true model parameters. The performance of UP-true can be regarded as an informal upper bound for UP-OSI. 


Further, we demonstrate that UP-OSI can also perform well when the model parameters of the testing environment are outside of the training range. This result is particularly interesting when comparing against UP-true. We found that UP-OSI unexpectedly outperforms the UP-true under dynamic models that were unseen during training.

All results presented in this work are simulated in PyDart2 \cite{pydart}, a python wrapper for DART \cite{DART}\footnote{Supplementary video at \textit{https://youtu.be/dwyuScnPNME}}, which is a multibody physics simulator supported by Gazebo. The simulation timestep is set to $0.002s$. Please see supplementary video for visualizing simulated motion sequences. For UP, we use a neural network with two hidden layers, comprised of 64 units in both hidden layers with tanh activation functions, followed by a linear fully connected final layer. For OSI, we use three hidden layers, with 256, 128, and 64 hidden units and tanh activation functions. We add a dropout layer for OSI after each hidden layer with a dropout rate of 0.1.


The learning process for UP takes 500 iterations of TRPO updates. The amount of data collected during each iteration varies by the difficulty of the tasks. We run five iterations for training OSI. At each iteration we sample $30$ different $\boldsymbol{\mu}$ values and collect $5$ seconds for each $\boldsymbol{\mu}$. For all the examples shown here, we use a motion history length of three, which is the minimal length to learn a second-order dynamic model.

\subsection{Double inverted pendulum with unknown center of mass}

\begin{figure*}[t!]
\centering
\subfigure{\includegraphics[width=15cm]{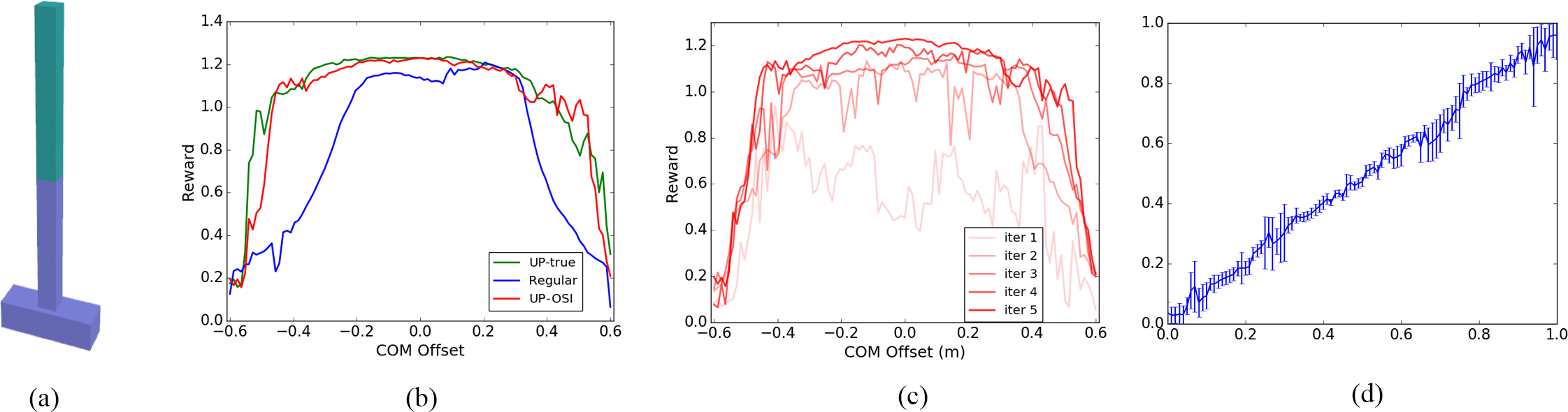}}
\caption{Results for the Double Inverted Pendulum Task. (a) Illustration of the task. (b) Performance of UP-true, "Regular" Controller and UP-OSI. The horizontal axis is the model parameter (center of mass), and the vertical axis is the performance. (c) The evolution of optimizing UP-OSI. The reward across the training range of $\mu$ increases iteratively. (d) Mean and standard deviation of the predicted model parameter. The x-axis indicates the true model parameters while the y-axis indicates the predicted ones by OSI. The model parameters have been normalized to be in $[-1, 1]$}
\label{fig:dpeninv_result}
\vspace{-3mm}
\end{figure*}


We begin with a classic motor control problem: balancing a double inverted pendulum. We define the reward function as,
\begin{equation}
r(\vc{x}) = -k_1(\sigma_1+\sigma_2)^2 - k_2|p_{cart}| + 10, \nonumber
\end{equation}
where $\sigma_1$ and $\sigma_2$ are angles of the two poles from the upright configuration, $p_{cart}$ is the position of the cart and $k_1$ and $k_2$ are the corresponding weights of the two terms. We normalize the angles to be in $[0, \pi]$ and use $k_1=10.0, k_2=1.0$ in our experiment. The length of the two poles are both $0.5m$. We terminate the simulation when $|p_{cart} \geq 5|$ or $(\sigma_1+\sigma_2) \geq 0.5\pi$.

The unknown model parameter for this problem is the center of mass of the lower pole, which has an unknown offset, $(\mu, 0.2\mu)$, from the geometric center. To ensure that the control policy would need to apply different strategies to balance the pendulum when different model parameters are given, we allow the offset to vary across a wide range: $\mu \in [-0.6m, 0.6m]$. Note that the purpose of the vertical offset ($0.2 \mu)$ is to break the symmetry of the problem to further increase the difficulty of control.



%

At each training iteration of UP, we collect $150,000$ samples using the physics simulator. Figure \ref{fig:dpeninv_result}(b) shows the normalized performance of the trained UP-OSI across different $\boldsymbol{\mu}$ values, comparing against UP-true (the informal upper bound). The performance of each $\boldsymbol{\mu}$ in Figure \ref{fig:dpeninv_result}(b) is the normalized average accumulated reward of $20$ rollouts starting from a randomly perturbed initial state. If the performance value is above $1.0$, the double inverted pendulum is able to balance. We also compare UP-OSI to a policy with conventional state input and control output but trained by data simulated from a range of model parameters (denoted as ``regular'' in Figure \ref{fig:dpeninv_result}(b)). The purpose of this comparison is to show that providing the model parameters as input to UP results in a more powerful control policy under a range of dynamic models, 

To demonstrate the learning process of OSI network, we plot the same reward-model parameter graph at each iteration of training. As shown in Figure \ref{fig:dpeninv_result}(c), the performance of UP-OSI improves over time and approaches the performance of UP-true. In Figure \ref{fig:dpeninv_result}(d), we plot the mean and the standard deviation for the model parameter identified by the trained OSI for each ground truth $\mu$ on x-axis. This shows that OSI is indeed able to identify the model parameter in this task.

\subsection{Manipulator with unknown object mass}

\begin{figure*}[t!]
\centering
\subfigure{\includegraphics[width=12cm]{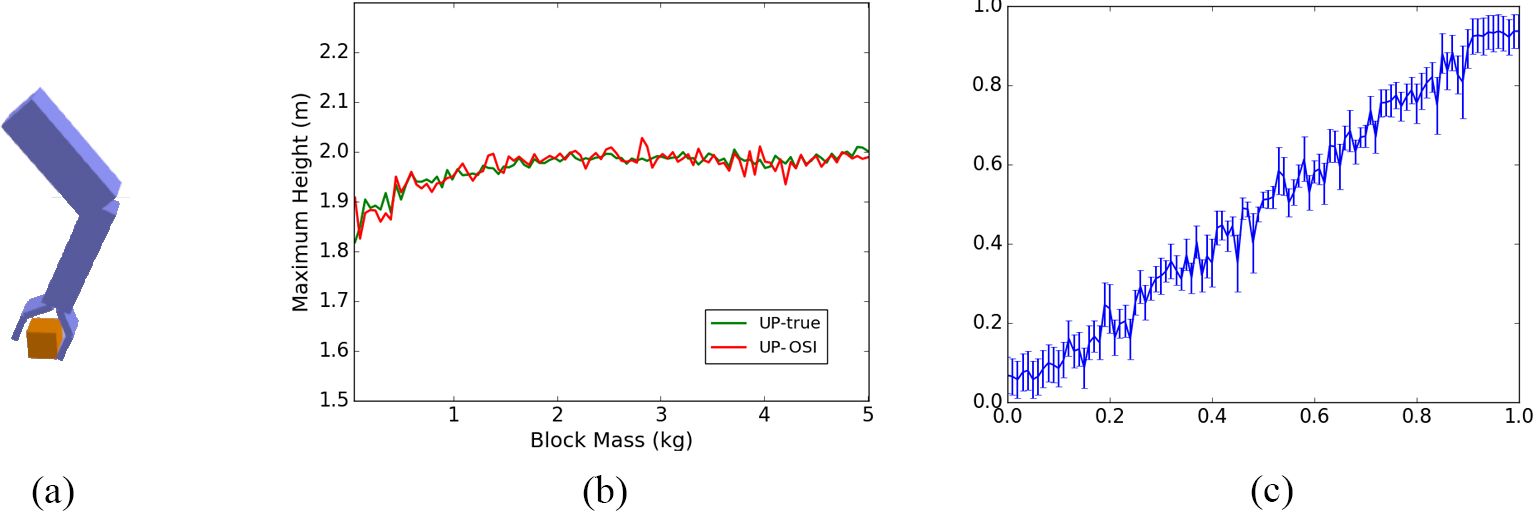}}
\caption{Results for the Robot Arm object throwing task. (a) Illustration of the Task. (b) Performance of UP-true and UP-OSI. The horizontal axis is the model parameter (mass of the object), and the vertical axis is the performance (maximum height of the object). (c) Mean and standard deviation of the predicted model parameter. The x-axis indicates the true model parameters while the y-axis indicates the predicted ones by OSI. The model parameters have been normalized to be in $[-1, 1]$.}
\label{fig:gripper_result}
\vspace{-3mm}
\end{figure*}

In this example, we train a robot arm to grab a block and throw it to a certain height but not beyond. The arm is initially pointing down and the block is in the air near the gripper of the arm. Similar motor skill can be observed in the serving of a tennis ball. The state $\vc{x}$ includes the joint position $\vc{q}$, joint velocity $\dot{\vc{q}}$ of the robot arm, and the position of the block $\vc{p}_{block}$. The reward function is defined as:
\begin{align*}
&r(\vc{x},\vc{u})=-k_1 r_{h} - k_2 ||\vc{u}||^2 - k_3 ||\dot{q}||^2 + 35\\ 
&r_{h} = 
\begin{cases}
    h_{target}-h_{block},& \text{if } h_{block}\leq h_{target}\\        0,         & \text{otherwise}
\end{cases},
\end{align*}
where $k_1=10, k_2=1e-5, k_3=1e-3,h_{target} = 2m$ and $h_{block}$ is the height of the block. We terminate the rollout when the box falls below $-0.2m$ or when the block is more than $0.8m$ away horizontally. By giving zero reward beyond $h_{target}$, we encourage the robot arm to throw the block in a way that it has low velocity when it reaches $h_{target}$, such that it can stay in the high reward region as long as possible. The unknown model parameters is the mass of the block. The robot needs to infer the weight of the block and use the right amount of effort to throw it up to the right height. 

During the training of UP, we collect $50,000$ samples for each iteration. The performance of UP-OSI and UP-true is plotted in Figure\ref{fig:gripper_result}(b). We measure the performance by the highest point reached by the block. The closer to $h_{target}=2m$, the better the performance. We also plot the mean and standard deviation of the predicted block mass throughout the test, as shown in Figure\ref{fig:gripper_result}(c). We observe similar trends as in the double inverted pendulum task, which shows that OSI also learns to identify the model parameter for this task.

\subsection{Hopper with unknown friction coefficient}

%

Correctly identifying contact information is crucial to many locomotion tasks. In this example, we demonstrate that our method can be applied to identify the friction coefficient at the contact point in an online fashion. The task is to control a single leg robot in 2D, the Hopper, to hop forward as fast as possible without falling. The reward is defined as
\begin{equation}
r(\vc{x}, \vc{u}) = k_1 \dot{\vc{x}} - k_2 ||\vc{u}||^2 + 3.0, \nonumber
\end{equation}
where $k_1=1, k_2=0.002$ for our experiments. The unknown model parameters is the friction coefficient with the range $\boldsymbol{\mu} \in [0.3, 1.0]$. We plot the maximum distance traveled by the hopper before the termination criteria is satisfied (the hopper falls or the maximum length of the rollout is reached) instead of the reward value to better visualize the performance of the hopper. The input to UP include the joint position of the hopper $\vc{q}$, the joint velocity $\dot{\vc{q}}$ and the friction coefficient $\mu$ between the foot and the ground. Note that we don't use position in the forward direction in the input state, because it is not directly related to the task.

\begin{figure*}[t!]
\centering
\subfigure{\includegraphics[width=15cm]{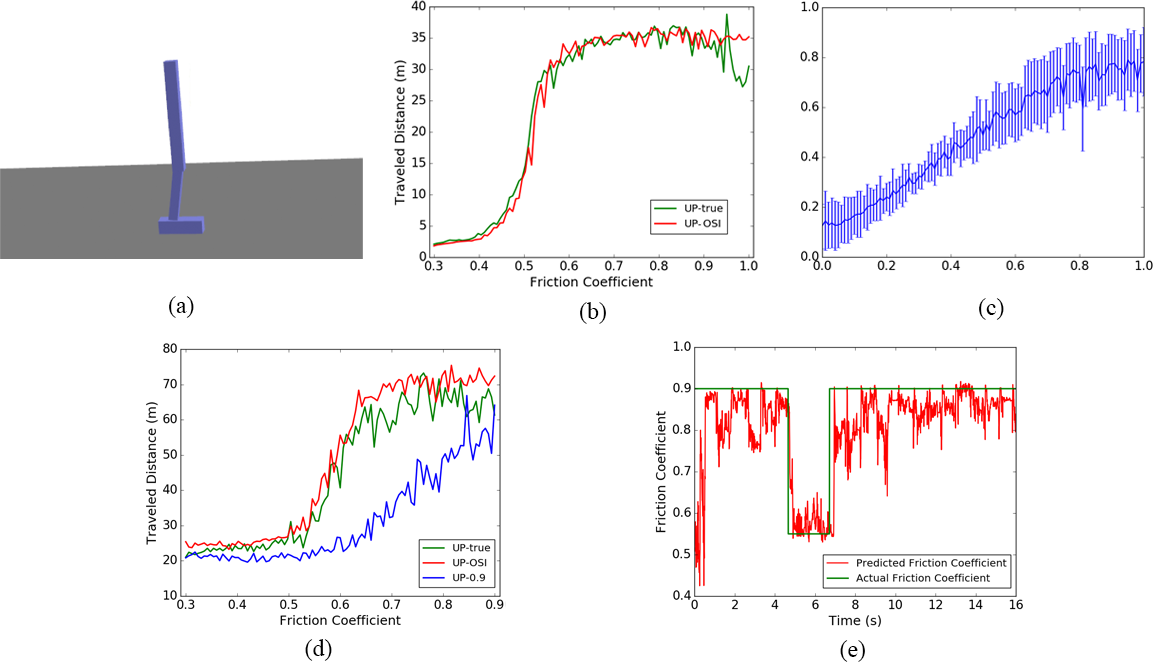}}
\caption{Results for the Hopper task. (a) Illustration of the task. (b) Performance of UP-true and UP-OSI. The horizontal axis is the model parameter (friction coefficient), and the vertical axis is the performance (maximum distance traveled in $1,000$ simulation steps). (c) Mean and standard deviation of the predicted model parameter. The x-axis indicates the true model parameters while the y-axis indicates the predicted ones by OSI. The model parameters have been normalized to be in $[-1, 1]$. (d) Performance of varying contact friction test. We tested UP-true, UP-OSI and UP with input friction coefficient fixed at $0.9$ for $2,000$ simulation steps. (e) OSI-predicted and actual friction coefficient in varying contact friction test.}
\label{fig:hopper_result}
\vspace{-3mm}
\end{figure*}

We use $75,000$ samples each iteration during the training of UP. Figure~\ref{fig:hopper_result}(b) shows the performance of UP-OSI compared to UP-true. Due to the difficulty of the task, UP can only perform well around $\mu = [0.6, 1.0]$. However, being able to identify friction coefficients in this range, \ie between the coefficient for wood-concrete contact and that for rubber-concrete contact, is sufficient for most practical applications. Figure~\ref{fig:hopper_result}(c) shows the mean and standard deviation of the predicted model parameter during the test at each $\boldsymbol{\mu}$.


\subsection{Cart-pole swing-up with unknown pole length and unknown attached mass}

\begin{figure*}[t!]
\centering
\subfigure{\includegraphics[width=15cm]{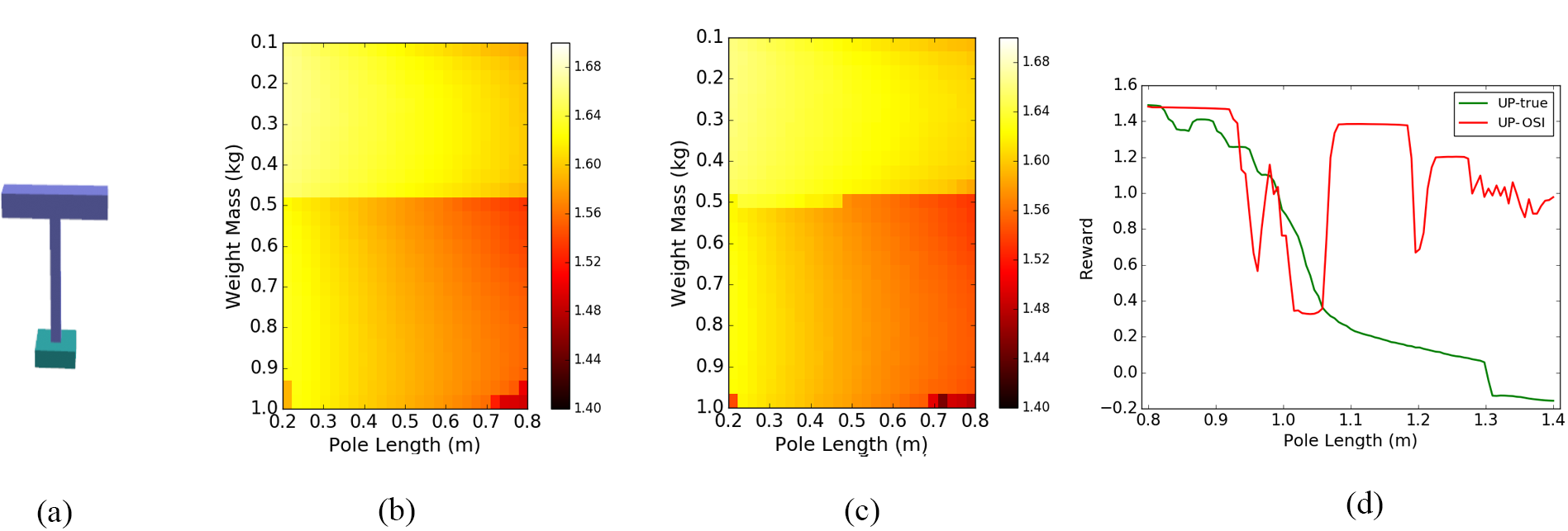}}
\caption{Results on the Cart-Pole Swing-Up task. (a) Illustration of the task. (b) Performance of UP-true visualized in the domain of pole length and weight mass. (c) Performance of UP-OSI visualized in the domain of pole length and weight mass. (d) Performance with model parameters exceeding the training range $100\%$.} 
\label{fig:swingup_result}
\vspace{-3mm}
\end{figure*}

To solve the classic cart-pole swing-up problem, the control policy needs to learn not only how to balance the pole, but also how to swing it up from a straight down position. Our experiment makes two modifications to increase the difficulty of the problem. First, we limit the force used by the cart to be within $[-40N, 40N]$. As such, the controller must swing the pole back and forth before it rises up. We also attach an additional mass to the tip of the pole to mimic the weight lifting task (Figure \ref{fig:swingup_result}(a)).

We use a variant of the reward function suggested by \cite{levine2015learning}: $r_\sigma=w\sigma^2+v \log(\sigma^2+a),$ where $\sigma$ is the angle of the pole. The first term encourages fast learning of swing-up motion and the second term encourages fast learning of balance. In our experiment, we set $w=1, v=1, a=0.1$. Similar to the double inverted pendulum task, we also add a term to encourage the cart to stay at the center of the track, $r_{cart} = |p_cart|$. Together, our reward function is defined as: 
\begin{equation}
r(\vc{x}) = - k_1 r_\sigma - k_2 r_{cart} + 10.0, \nonumber
\end{equation}
where $k_1=1.0, k_2=0.2$. In our setup, the pole has $0$ position when it is upright. During the simulation, we randomly initialize the position of the pole to be either $\pi$ or $-\pi$ with a small noise drawn from $\mathcal{N}(0, 0.005)$. We terminate the rollout when the pole rotates more than $4\pi$ from the initial position, or when the cart is more than $2m$ from the center.


The unknown model parameters in this example includes the additional mass attached to the top of the pole ($\boldsymbol{\mu}_{mass} \in [0.1kg, 1.0kg]$), and the length of the pole ($\boldsymbol{\mu}_{length} \in [0.2m, 0.8m]$). To closely compare with the cart-pole examples in \cite{heess2015memory}, where they control an inverted pendulum with varying pole length using a RNN with the whole history trajectory as input, we train OSI to estimate the velocity of the system, instead of directly giving the true velocity to the policy as part of the state. As such, the space of model parameters for this example is $\mathbb{R}^4$. At each iteration of UP training, we run $70,000$ samples. We normalize the resulting reward such that if a policy achieves averaged accumulated reward more than one, then it usually can swing up and balance the cart-pole system. Figure \ref{fig:swingup_result}(b) and (c) show that UP-OSI can achieve high reward for a range of unknown pole lengths, similar to the inverted pendulum result shown in \cite{heess2015memory}, but UP-OSI only requires three time steps of history as input. In addition, the mass attached to the tip is also an unknown that needs to be simultaneously identified with the pole length.

\subsection{Generalization to varying model parameter}

We run the trained UP-OSI for Hopper on a track with varying friction coefficients to test its generalizability. We create a track with friction coefficient $\boldsymbol{\mu}_{const}=0.9$ everywhere except for the region between $20m$ to $30m$. We then vary the friction coefficient $\boldsymbol{\mu}_{vary}$ in this region and plot the performance of the controller with regard to $\boldsymbol{\mu}_{vary}$. Figure~\ref{fig:hopper_result}(d) shows the performance of UP-OSI and the UP-true. Note that UP-true was given the ground truth friction coefficient at each time instance as if it has a perfect contact-friction sensor on the foot, where UP-OSI needs to identify this information based on the recent history of the motion. The results show that UP-OSI can achieve comparable and sometimes better performance than UP-true. We also test the performance of UP with fixed input $\boldsymbol{\mu}=0.9$, \ie the hopper isn't aware of the change in friction coefficient, which is shown as the blue curve in Figure~\ref{fig:hopper_result}(d). The worse performance shows that it is crucial to detect the varying friction coefficient in order to success in this task.

In addition, we plot the friction coefficient predicted by OSI over time for one specific $\boldsymbol{\mu}_{vary}=0.55$, as is shown in Figure~\ref{fig:hopper_result}(e). We can see that OSI can identify the changes in model parameter during the task. Note that we did not provide any training examples with temporally-varying $\mu$ when training either UP-true or UP-OSI networks.

\subsection{Generalization beyond training range}
Another way to evaluate the generalizability of the policy is to test it with model parameters that were not seen during the training phase. We perform such test on the cart-pole swing up problem with a pole length range of $[0.8 m, 1.4 m]$ and the attached mass range of $[1.0 kg, 1.9 kg]$. We linearly couple the two unknown parameters together, such that when pole length is $0.8m$ the attached mass has $1.0 kg$ and when pole length is $1.4m$ the attached mass has $1.9 kg$. This is $100\%$ beyond the original training range ($[0.2 m, 0.8 m]$ and $[0.1 kg, 1.0 kg]$). We test both UP-true and UP-OSI with this extended range, which is shown in Figure~\ref{fig:swingup_result}(d). The result shows that UP-OSI, can work for a large range of unknown pole length and attached mass with only position information as input. More interestingly, UP-OSI significantly outperforms UP-true in this range of $\boldsymbol{\mu}$ unseen during training.

\section{Discussion}
While UP-OSI has demonstrated a wide range of success, we recognize a few limitations that require further investigation. First, identifying high-dimensional model parameters remains an unsolved challenge. Although we demonstrated that OSI can perform well for model parameters in $\mathbb{R}^4$ (the cart-pole example), more rigorous analysis is required to evaluate the sample-efficiency of UP-OSI for high-dimensional model parameter space. In all our experiments, we train OSI with five iterations and the resulting UP-OSI can achieve similar performance as the baseline. However, a theoretical upper bound and convergence conditions are not established in this work. Further, our current implementation assumes both the policy and the dynamic model are deterministic, but UP-OSI can easily be extend to a stochastic formulation. 

This work primarily focuses on identifying model parameters unrelated to uncertainty in sensors and actuators partly because many of these issues have been addressed in the literature of robust control. One important source of uncertainty in a dynamic model is  latency. The nature of latency might complicate the scheme of history queue used in OSI, but it is nevertheless an important and challenging problem to tackle. 

An alternative approach to UP-OSI is to train an ``end-to-end'' control policy that takes as input a sequence of motion history and directly outputs the optimal control, rather than explicitly decoupling the process into system identification and control. Indeed, previous work \cite{heess2015memory} has shown that a recurrent network can learn to control a dynamic system with unknown model parameters. We chose to design a control policy that explicitly takes in the model parameters as input, because this additional information in the input allows the policy to specialize for each model parameter, which can potentially improve the performance. In addition, we conjecture that decoupling a large network trained by reinforcement learning into two moderate-sized networks, one of which is trained by supervised learning, might reduce the learning time and improve the sample-efficiency.

We demonstrated that UP-OSI can be extended to operate under the model parameters outside of the training range. The more surprising result is that, in these extrapolated situations, UP-OSI actually outperforms the baseline. While further investigation is needed, one possible explanation is that OSI learned how to modulate the model parameters over time, essentially switching between existing policies, to achieve unseen tasks. Since UP-OSI has the ability to handle $\boldsymbol{\mu}$ outside of the training range, a follow-up question for future work is whether UP-OSI can handle variations in other model parameters outside of the space of $\boldsymbol{\mu}$.

For near future, we would like to train UP-OSI using one simulator but test it on a different one. If successful, our next step will be to test UP-OSI on real-world robotic platforms. One potential issue of our method is that the real-world dynamics cannot be well represented by the chosen parameterization of the dynamic models. However, UP-OSI allows the dynamic model to be time-varying, resulting in a more expressive parametric model to match the real-world dynamics. Essentially, the algorithm modifies its estimate of the dynamic model at every time instance based on recent real-world observations. This rapid adaptation of dynamic model allows the control policy to make prompt correction as it is also time-varying.




\section{Conclusion}

We have introduced a new approach to creating motion policies that are formed by coupling a Universal Policy together with an On-line System Identification model.  A key aspect of such policies is that they are robust under a wide variety of dynamic models, and indeed, they determine the dynamic model parameters on-the-fly.  This approach was created with the goal of developing control policies for real-world tasks by performing extensive off-line training using physical simulation and deep reinforcement learning.  While we have yet to transfer such policies to real robots, their performance gives several promising indications.  First, the UP-OSI control policies give almost the same performance as the UP-true baseline, while having to deduce the unknown dynamic models on their own.  Second, UP-OSI can perform \emph{better} than the Universal Policy alone, in the case where the dynamic model is outside of the training range.  Finally, by constantly estimating the dynamic model at every time-step, such control policies can adapt to a changing environment such as locomotion over a surface with varying friction.




\bibliographystyle{plainnat}
\bibliography{references}

\begin{thebibliography}{34}
\providecommand{\natexlab}[1]{#1}
\providecommand{\url}[1]{\texttt{#1}}
\expandafter\ifx\csname urlstyle\endcsname\relax
  \providecommand{\doi}[1]{doi: #1}\else
  \providecommand{\doi}{doi: \begingroup \urlstyle{rm}\Url}\fi

\bibitem[DAR()]{DART}
D{A}{R}{T}: {D}ynamic {A}nimation and {R}obotics {T}oolkit".
\newblock URL \url{http://dartsim.github.io/}.

\bibitem[pyd()]{pydart}
Pydart2.
\newblock URL \url{https://github.com/sehoonha/pydart2}.

\bibitem[Abbeel and Ng(2005)]{Abbeel:2005}
Pieter Abbeel and Andrew~Y. Ng.
\newblock \href{http://dl.acm.org/citation.cfm?id=1102352}{Exploration and
  Apprenticeship Learning in Reinforcement Learning}.
\newblock In \emph{International Conference on Machine Learning}, pages 1--8,
  2005.

\bibitem[Abbeel et~al.(2006)Abbeel, Quigley, and Ng]{Abbeel:2006}
Pieter Abbeel, Morgan Quigley, and Andrew~Y. Ng.
\newblock \href{http://dl.acm.org/citation.cfm?id=1143845}{Using Inaccurate
  Models in Reinforcement Learning}.
\newblock In \emph{Proceedings of the 23rd International Conference on Machine
  Learning}, ICML '06, pages 1--8, 2006.

\bibitem[Bongard and Lipson(2005)]{journals/tec/BongardL05}
Josh~C. Bongard and Hod Lipson.
\newblock \href{http://dx.doi.org/10.1109/TEVC.2005.850293}{Nonlinear System
  Identification Using Coevolution of Models and Tests}.
\newblock \emph{IEEE Trans. Evolutionary Computation}, 9\penalty0 (4):\penalty0
  361--384, 2005.
\newblock URL \url{http://dx.doi.org/10.1109/TEVC.2005.850293}.

\bibitem[Christiano et~al.(2016)Christiano, Shah, Mordatch, Schneider,
  Blackwell, Tobin, Abbeel, and Zaremba]{christiano2016transfer}
Paul Christiano, Zain Shah, Igor Mordatch, Jonas Schneider, Trevor Blackwell,
  Joshua Tobin, Pieter Abbeel, and Wojciech Zaremba.
\newblock \href{https://arxiv.org/abs/1610.03518}{Transfer from Simulation to
  Real World through Learning Deep Inverse Dynamics Model}.
\newblock \emph{arXiv preprint arXiv:1610.03518}, 2016.

\bibitem[Da~Silva et~al.(2012)Da~Silva, Konidaris, and Barto]{da2012learning}
Bruno Da~Silva, George Konidaris, and Andrew Barto.
\newblock \href{https://arxiv.org/abs/1206.6398}{Learning parameterized
  skills}.
\newblock \emph{arXiv preprint arXiv:1206.6398}, 2012.

\bibitem[Deisenroth and Rasmussen(2011)]{deisenroth2011}
Marc Deisenroth and Carl~E Rasmussen.
\newblock \href{http://www.icml-2011.org/papers/323_icmlpaper.pdf}{PILCO: A
  model-based and data-efficient approach to policy search}.
\newblock In \emph{Proceedings of the 28th International Conference on machine
  learning (ICML-11)}, pages 465--472, 2011.

\bibitem[Gevers(2006)]{gevers2006}
Michel Gevers.
\newblock System identification without lennart ljung: what would have been
  different?
\newblock 2006.

\bibitem[Gu et~al.(2016)Gu, Lillicrap, Ghahramani, Turner, and Levine]{gu2016q}
Shixiang Gu, Timothy Lillicrap, Zoubin Ghahramani, Richard~E Turner, and Sergey
  Levine.
\newblock \href{https://arxiv.org/abs/1611.02247}{Q-Prop: Sample-Efficient
  Policy Gradient with An Off-Policy Critic}.
\newblock \emph{arXiv preprint arXiv:1611.02247}, 2016.

\bibitem[Ha and Yamane(2015)]{HA:2015}
Sehoon Ha and Katsu Yamane.
\newblock \href{http://ieeexplore.ieee.org/document/7139552/}{Reducing Hardware
  Experiments for Model Learning and Policy Optimization}.
\newblock \emph{IROS}, 2015.

\bibitem[Heess et~al.(2015)Heess, Hunt, Lillicrap, and Silver]{heess2015memory}
Nicolas Heess, Jonathan~J Hunt, Timothy~P Lillicrap, and David Silver.
\newblock \href{https://arxiv.org/abs/1512.04455}{Memory-based control with
  recurrent neural networks}.
\newblock \emph{arXiv preprint arXiv:1512.04455}, 2015.

\bibitem[James and Johns(2016)]{james20163d}
Stephen James and Edward Johns.
\newblock \href{https://arxiv.org/abs/1609.03759}{3D Simulation for Robot Arm
  Control with Deep Q-Learning}.
\newblock \emph{arXiv preprint arXiv:1609.03759}, 2016.

\bibitem[Koos et~al.(2010)Koos, Mouret, and Doncieux]{conf/gecco/KoosMD10}
Sylvain Koos, Jean-Baptiste Mouret, and St{\'e}phane Doncieux.
\newblock Crossing the reality gap in evolutionary robotics by promoting
  transferable controllers.
\newblock In \emph{Genetic and Evolutionary Computation Conference}. ACM, 2010.
\newblock ISBN 978-1-4503-0072-8.
\newblock URL \url{http://doi.acm.org/10.1145/1830483.1830505}.

\bibitem[Levine et~al.(2015)Levine, Wagener, and Abbeel]{levine2015learning}
Sergey Levine, Nolan Wagener, and Pieter Abbeel.
\newblock \href{https://arxiv.org/abs/1501.05611}{Learning contact-rich
  manipulation skills with guided policy search}.
\newblock In \emph{Robotics and Automation (ICRA), 2015 IEEE International
  Conference on}, pages 156--163. IEEE, 2015.

\bibitem[Levine et~al.(2016{\natexlab{a}})Levine, Finn, Darrell, and
  Abbeel]{DBLP:journals/corr/LevineFDA15}
Sergey Levine, Chelsea Finn, Trevor Darrell, and Pieter Abbeel.
\newblock \href{http://dl.acm.org/citation.cfm?id=2946645.2946684}{End-to-End
  Training of Deep Visuomotor Policies}.
\newblock \emph{Journal of Machine Learning Research}, 17\penalty0
  (39):\penalty0 1--40, 2016{\natexlab{a}}.

\bibitem[Levine et~al.(2016{\natexlab{b}})Levine, Pastor, Krizhevsky, and
  Quillen]{DBLP:journals/corr/LevinePKQ16}
Sergey Levine, Peter Pastor, Alex Krizhevsky, and Deirdre Quillen.
\newblock \href{https://arxiv.org/abs/1603.02199}{Learning Hand-Eye
  Coordination for Robotic Grasping with Deep Learning and Large-Scale Data
  Collection}.
\newblock \emph{CoRR}, abs/1603.02199, 2016{\natexlab{b}}.
\newblock URL \url{http://arxiv.org/abs/1603.02199}.

\bibitem[Lillicrap et~al.(2015)Lillicrap, Hunt, Pritzel, Heess, Erez, Tassa,
  Silver, and Wierstra]{lillicrap2015continuous}
Timothy~P Lillicrap, Jonathan~J Hunt, Alexander Pritzel, Nicolas Heess, Tom
  Erez, Yuval Tassa, David Silver, and Daan Wierstra.
\newblock \href{https://arxiv.org/abs/1509.02971}{Continuous control with deep
  reinforcement learning}.
\newblock \emph{arXiv preprint arXiv:1509.02971}, 2015.

\bibitem[Mnih et~al.(2015)Mnih, Kavukcuoglu, Silver, Rusu, Veness, Bellemare,
  Graves, Riedmiller, Fidjeland, Ostrovski, et~al.]{mnih2015human}
Volodymyr Mnih, Koray Kavukcuoglu, David Silver, Andrei~A Rusu, Joel Veness,
  Marc~G Bellemare, Alex Graves, Martin Riedmiller, Andreas~K Fidjeland, Georg
  Ostrovski, et~al.
\newblock
  \href{http://www.nature.com/nature/journal/v518/n7540/full/nature14236.html}{Human-level
  control through deep reinforcement learning}.
\newblock \emph{Nature}, 518\penalty0 (7540):\penalty0 529--533, 2015.

\bibitem[Mnih et~al.(2016)Mnih, Badia, Mirza, Graves, Lillicrap, Harley,
  Silver, and Kavukcuoglu]{mnih2016asynchronous}
Volodymyr Mnih, Adria~Puigdomenech Badia, Mehdi Mirza, Alex Graves, Timothy~P
  Lillicrap, Tim Harley, David Silver, and Koray Kavukcuoglu.
\newblock \href{https://arxiv.org/abs/1602.01783}{Asynchronous methods for deep
  reinforcement learning}.
\newblock In \emph{International Conference on Machine Learning}, 2016.

\bibitem[Mordatch et~al.(2015)Mordatch, Lowrey, and Todorov]{Mordatch:2015}
Igor Mordatch, Kendall Lowrey, and Emanuel Todorov.
\newblock \href{http://ieeexplore.ieee.org/document/7354126/}{Ensemble-CIO:
  Full-body dynamic motion planning that transfers to physical humanoids}.
\newblock In \emph{IEEE/RSJ International Conference on Intelligent Robots and
  Systems}, 2015.

\bibitem[Nolfi and Floreano(2000)]{realityGap}
S.~Nolfi and D.~Floreano.
\newblock \emph{Evolutionary Robotics: The Biology, Intelligence, and
  Technology}.
\newblock MIT Press (Cambridge, MA), 2000.

\bibitem[Peng et~al.(2016)Peng, Berseth, and van~de Panne]{2016-TOG-deepRL}
Xue~Bin Peng, Glen Berseth, and Michiel van~de Panne.
\newblock \href{http://dl.acm.org/citation.cfm?id=2925881}{Terrain-Adaptive
  Locomotion Skills Using Deep Reinforcement Learning}.
\newblock \emph{ACM Transactions on Graphics}, 35\penalty0 (4), 2016.

\bibitem[Pinto and Gupta(2016)]{pinto2016supersizing}
Lerrel Pinto and Abhinav Gupta.
\newblock \href{http://ieeexplore.ieee.org/document/7487517} {Supersizing
  self-supervision: Learning to grasp from 50k tries and 700 robot hours}.
\newblock In \emph{IEEE International Conference on Robotics and Automation
  (ICRA)}, pages 3406--3413. IEEE, 2016.

\bibitem[Punjani and Abbeel(2015)]{punjani2015deep}
Ali Punjani and Pieter Abbeel.
\newblock \href{http://ieeexplore.ieee.org/document/7139643/}{Deep learning
  helicopter dynamics models}.
\newblock In \emph{IEEE International Conference on Robotics and Automation
  (ICRA)}, pages 3223--3230. IEEE, 2015.

\bibitem[Rajeswaran et~al.(2016)Rajeswaran, Ghotra, Levine, and
  Ravindran]{rajeswaran2016epopt}
Aravind Rajeswaran, Sarvjeet Ghotra, Sergey Levine, and Balaraman Ravindran.
\newblock \href{https://arxiv.org/abs/1610.01283}{EPOpt: Learning Robust Neural
  Network Policies Using Model Ensembles}.
\newblock \emph{arXiv preprint arXiv:1610.01283}, 2016.

\bibitem[Ross and Bagnell(2012)]{Ross12agnosticsystem}
Stephane Ross and J.~Andrew Bagnell.
\newblock \href{https://arxiv.org/abs/1203.1007}{Agnostic system identification
  for model-based reinforcement learning}.
\newblock In \emph{In ICML}, 2012.

\bibitem[Rusu et~al.(2016)Rusu, Vecerik, Roth{\"o}rl, Heess, Pascanu, and
  Hadsell]{rusu2016sim}
Andrei~A Rusu, Matej Vecerik, Thomas Roth{\"o}rl, Nicolas Heess, Razvan
  Pascanu, and Raia Hadsell.
\newblock \href{https://arxiv.org/abs/1610.04286}{Sim-to-real robot learning
  from pixels with progressive nets}.
\newblock \emph{arXiv preprint arXiv:1610.04286}, 2016.

\bibitem[Schulman et~al.(2015{\natexlab{a}})Schulman, Levine, Moritz, Jordan,
  and Abbeel]{schulman2015trust}
John Schulman, Sergey Levine, Philipp Moritz, Michael~I Jordan, and Pieter
  Abbeel.
\newblock \href{https://arxiv.org/abs/1502.05477}{Trust region policy
  optimization}.
\newblock \emph{CoRR, abs/1502.05477}, 2015{\natexlab{a}}.

\bibitem[Schulman et~al.(2015{\natexlab{b}})Schulman, Moritz, Levine, Jordan,
  and Abbeel]{schulman2015high}
John Schulman, Philipp Moritz, Sergey Levine, Michael Jordan, and Pieter
  Abbeel.
\newblock \href{https://arxiv.org/abs/1506.02438}{High-dimensional continuous
  control using generalized advantage estimation}.
\newblock \emph{arXiv preprint arXiv:1506.02438}, 2015{\natexlab{b}}.

\bibitem[Silver et~al.(2014)Silver, Lever, Heess, Degris, Wierstra, and
  Riedmiller]{Silver2014DeterministicPG}
David Silver, Guy Lever, Nicolas Heess, Thomas Degris, Daan Wierstra, and
  Martin~A. Riedmiller.
\newblock
  \href{http://jmlr.org/proceedings/papers/v32/silver14.html}{Deterministic
  Policy Gradient Algorithms}.
\newblock In \emph{ICML}, 2014.

\bibitem[Stulp et~al.(2013)Stulp, Raiola, Hoarau, Ivaldi, and
  Sigaud]{stulp2013learning}
Freek Stulp, Gennaro Raiola, Antoine Hoarau, Serena Ivaldi, and Olivier Sigaud.
\newblock \href{http://ieeexplore.ieee.org/document/7030008/}{Learning compact
  parameterized skills with a single regression}.
\newblock In \emph{Humanoid Robots (Humanoids), 2013 13th IEEE-RAS
  International Conference on}, pages 417--422. IEEE, 2013.

\bibitem[Szita et~al.(2002)Szita, Tak{\'a}cs, and
  L{\"o}rincz]{szita2002varepsilon}
Istv{\'a}n Szita, B{\'a}lint Tak{\'a}cs, and Andr{\'a}s L{\"o}rincz.
\newblock
  \href{http://www.jmlr.org/papers/v3/szita02a.html}{$\varepsilon$-MDPs:
  Learning in varying environments}.
\newblock \emph{Journal of Machine Learning Research}, 3\penalty0
  (Aug):\penalty0 145--174, 2002.

\bibitem[Zagal et~al.(2004)Zagal, {Ruiz-del-Solar}, and Vallejos]{zagal2004}
Juan~Cristobal Zagal, Javier {Ruiz-del-Solar}, and Paul Vallejos.
\newblock \href{http://dl.acm.org/citation.cfm?id=1830505}{{B}ack-to-{R}eality:
  Crossing the Reality Gap in Evolutionary Robotics}.
\newblock In \emph{IAV 2004: Proceedings 5th IFAC Symposium on Intelligent
  Autonomous Vehicles}. Elsevier Science Publishers~B.V., 2004.
\newblock ISBN 008-044237-4.

\end{thebibliography}

\end{document}